\documentclass[letterpaper, 10 pt, conference]{ieeeconf}
\IEEEoverridecommandlockouts
\usepackage{amsmath} 
\usepackage{amssymb}
\usepackage{amsthm}
\usepackage{upgreek}
\newtheoremstyle{mystyle}%
  {\topsep}%
  {\topsep}%
  {\normalfont}%
  {}%
  {\bfseries}%
  {}%
  {.5em}%
  {}%
\theoremstyle{plain}

\usepackage{mathtools}
\usepackage{siunitx}
\usepackage{algorithm}
\usepackage{algpseudocode}
\usepackage{lipsum}
\usepackage[short,nocomma]{optidef}
\usepackage{booktabs}
\usepackage{graphicx}
\usepackage{tikz}
\usetikzlibrary{arrows.meta,calc}
\usepackage{pgfplots}
\usepgfplotslibrary{fillbetween,groupplots}
\pgfplotsset{compat=1.18}

\title{\LARGE \bf
Exploiting Hierarchical Controller Structure in Contextual Parameter Learning for Humanoid Loco-Manipulation
}

\author{Sebastian Hirt$^{1}$, Lukas Theiner$^{1}$, Jan Peters$^{2,3,4}$, and Rolf Findeisen$^{1}$%
\thanks{$^{1}$Control and Cyber-Physical Systems Laboratory (CCPS), TU Darmstadt,
{\tt\small \{sebastian.hirt, rolf.findeisen\}@iat.tu-darmstadt.de}}%
\thanks{$^{2}$Intelligent Autonomous Systems Laboratory, TU Darmstadt}%
\thanks{$^{3}$German Research Center for AI (DFKI), Research Department SAIROL}%
\thanks{$^{4}$Hessian Centre for Artificial Intelligence}%
}

\newcommand\copyrighttext{%
  \footnotesize This work has been submitted to the IEEE for possible publication. Copyright may be transferred without notice, after which this version may no longer be accessible.}

\newcommand\copyrightnotice{%
\begin{tikzpicture}[remember picture,overlay]
\node[anchor=south,yshift=5pt] at (current page.south) {\fbox{\parbox{\dimexpr0.8\textwidth-\fboxsep-\fboxrule\relax}{\copyrighttext}}};
\end{tikzpicture}%
}

\begin{document}

\maketitle
\thispagestyle{empty}
\pagestyle{empty}

\copyrightnotice

\begin{abstract}
Hierarchical control architectures are widely used to decompose complex control problems into interacting control levels and are particularly important in robotics, where planning, whole-body motion, and lower-level control must be coordinated across different levels of abstraction and time scales. Their overall closed-loop performance, however, depends strongly on parameters distributed across the hierarchy, such that tuning controllers on different levels independently may neglect relevant cross-layer interactions. We propose a contextual Bayesian optimization framework for joint parameter learning in hierarchical control systems. Rather than modeling closed-loop performance only as a scalar black-box function, we retain separate observations of task performance, realization quality, and control effort. A correlated multi-output Gaussian process models these performance components, while their known aggregation into the overall closed-loop objective is evaluated analytically. The formulation exploits three complementary consequences of hierarchical control: informative performance quantities exposed by the hierarchy, coupling between parameters of different controller levels, and variations of these relations with operating conditions.  We evaluate the approach for humanoid loco-manipulation, jointly tuning a centroidal predictive controller and a whole-body controller for physical box pushing under varying box mass.
The proposed method achieves the lowest mean empirical regret during both training and adaptation among the considered baselines.
\end{abstract}

\begin{figure}[t]
\centering
\definecolor{figSurrogate}{HTML}{477FB5}
\definecolor{figBO}{HTML}{315F8F}
\definecolor{figParam}{HTML}{82B2D7}
\definecolor{figUpper}{HTML}{85939A}
\definecolor{figLower}{HTML}{A0ADB3}
\definecolor{figPlant}{HTML}{BBC3C7}
\definecolor{figLine}{HTML}{3D769E}
\definecolor{figInk}{HTML}{22313D}
\definecolor{figAccent}{HTML}{C47C31}
\begin{tikzpicture}[
font=\footnotesize,
>=Latex,
arr/.style={
->,
line width=0.8pt,
draw=black!70
},
bluearr/.style={
->,
line width=0.9pt,
draw=figLine
},
ctxarr/.style={
->,
line width=0.9pt,
draw=figAccent
},
ctxline/.style={
line width=0.9pt,
draw=figAccent
},
attrib/.style={
->,
densely dotted,
line width=0.9pt,
draw=gray!70
},
ctrl/.style={
draw=none,
text=figInk,
minimum width=19mm,
minimum height=10mm,
align=center
},
plant/.style={
draw=none,
fill=figPlant,
text=figInk,
minimum width=19mm,
minimum height=10mm,
align=center
},
param/.style={
draw=none,
fill=figParam,
text=figInk,
minimum width=37mm,
minimum height=11mm,
align=center
},
surrogate/.style={
draw=none,
fill=figBO,
text=white,
minimum width=60mm,
minimum height=18mm,
align=center
},
bo/.style={
draw=none,
fill=figSurrogate,
text=white,
minimum width=60mm,
minimum height=14mm,
align=center
}
]
\node[ctrl, fill=figUpper] (upper) at (0,0) {
\textbf{Upper}\\[-0.5mm]
controller
};
\node[ctrl, fill=figLower] (lower) at (2.25,0) {
\textbf{Lower}\\[-0.5mm]
controller
};
\node[plant] (plant) at (4.50,0) {
\textbf{Physical}\\[-0.5mm]
system
};
\node[param] (param) at (1.125,1.55) {
\textbf{Joint parameters}\\[-0.2mm]
$\displaystyle
\theta=
\begin{bmatrix}
\theta_1^{\top} &
\theta_2^{\top}
\end{bmatrix}^{\top}$
};
\node[
text=figAccent,
align=center,
font=\footnotesize\bfseries
] (context) at (4.50,1.55) {
(3) Context $\chi$
};
\draw[bluearr]
([xshift=-6mm]param.south)
-- (upper.north);
\draw[bluearr]
([xshift=6mm]param.south)
-- (lower.north);
\draw[arr]
(upper.east)
-- (lower.west);
\draw[arr]
(lower.east)
-- (plant.west);
\draw[ctxarr]
(context.south)
-- (plant.north);
\node[surrogate] (surrogate) at (2.25,-2.10) {
\textbf{(1) Hierarchy-informed surrogate}\\[0.65mm]
$\mathbf J=
[\,J_{\mathrm{T}}\;\;J_{\mathrm{R}}\;\;J_{\mathrm{E}}\,]^\top$\\[0.45mm]
$\widehat{\mathbf J}^{(n)}(\theta,\chi)
\sim
\mathcal N(\boldsymbol{\mu}^{(n)},\Sigma^{(n)}),$\
$\widehat J^{(n)}=a^\top\widehat{\mathbf J}^{(n)}$
};
\draw[attrib]
(1.125,-0.05)
-- (1.125,-1.18);
\draw[attrib]
(3.375,-0.05)
-- (3.375,-1.18);
\draw[attrib]
(4.50,-0.52)
-- (4.50,-1.18);
\node[
text=black,
font=\scriptsize\bfseries,
fill=white,
inner sep=0.7pt
] at (1.125,-0.7) {
$J_{\mathrm{R}}$
};
\node[
text=black,
font=\scriptsize\bfseries,
fill=white,
inner sep=0.7pt
] at (3.375,-0.7) {
$J_{\mathrm{E}}$
};
\node[
text=black,
font=\scriptsize\bfseries,
fill=white,
inner sep=0.7pt
] at (4.50,-0.7) {
$J_{\mathrm{T}}$
};
\node[bo] (bo) at (2.25,-4.05) {
\textbf{(2) Cross-layer joint}\\[-0.2mm]
\textbf{Bayesian optimization}\\[0.7mm]
$\displaystyle
\theta^{(n+1)}
=
\arg\max_{\theta\in\Theta}
\alpha^{(n)}(\zeta(\theta,\chi))$
};
\draw[arr]
(surrogate.south)
-- (bo.north);
\coordinate (ctxTop) at (5.90,1.55);
\coordinate (ctxModel) at (5.90,-2.10);
\coordinate (ctxBO) at (5.90,-4.05);
\draw[ctxline]
(context.east)
-- (ctxTop)
-- (ctxModel)
-- (ctxBO);
\draw[ctxarr]
(ctxModel)
-- (surrogate.east);
\draw[ctxarr]
(ctxBO)
-- (bo.east);
\draw[bluearr]
(bo.west)
-- (-1.10,-4.05)
-- (-1.10,1.55)
-- (param.west);
\node[
anchor=south
] at (-1.25,1.66) {
$\theta^{(n+1)}$
};
\end{tikzpicture}%
\vspace{-3mm}
\caption{%
Overview of the proposed framework.
State-feedback paths within the controller hierarchy are omitted for clarity.
Dotted guides associate task ($J_{\mathrm{T}}$), realization ($J_{\mathrm{R}}$), and effort
($J_{\mathrm{E}}$) information with quantities exposed by the control hierarchy, while
all components may depend on the full parameter vector.
The hierarchy-informed surrogate provides the aggregated objective used
by the joint acquisition $\alpha^{(n)}$, while context dependence $\chi$ enables
transfer across operating conditions.
}
\vspace{-7mm}
\label{fig:method_overview}
\end{figure}

\vspace{-1mm}
\section{Introduction}
\vspace{-1mm}
Complex control systems are frequently organized in hierarchical
architectures, with individual layers operating at different levels
of abstraction and time scales~\cite{scattolini09,koegel25,baethge18}.
Such architectures are important in applications ranging from chemical
process control to aerospace systems and robotics.
More broadly, hierarchical and distributed control provide means
of managing complexity in interconnected cyberphysical
systems~\cite{lucia16}.
In robotics specifically, high-level planning, dynamic motion generation,
whole-body coordination, and lower-level control must often be
combined to realize complex physical behaviors. Humanoid and legged robots provide prominent examples: reduced-order models are commonly used to plan motion and contact interactions, while lower-level (whole-body) controllers realize these plans on the full articulated system subject to dynamics, contact, and actuation constraints~\cite{orin2013centroidal,kuindersma2016atlas,herzog2016momentum}. Similar hierarchical structures also arise in cascaded process control, trajectory tracking, and other multi-layer feedback systems.

While hierarchical decomposition makes complex online control problems tractable, it introduces a corresponding parameter-learning problem across interacting controller levels. The practical performance of each level depends not only on its own parameters, but also on how its outputs are interpreted and realized by the remaining hierarchy. Closed-loop parameter learning provides a means of adapting these parameters directly based on measured system performance~\cite{paulson2023tutorial,hirt2024safe}. Bayesian optimization (BO) is particularly suitable for this setting because it
targets expensive black-box objectives with a limited number of closed-loop
evaluations~\cite{akrour2017local_bo}. Beyond treating controller tuning as a generic black-box optimization problem, incorporating available prior knowledge and problem structure can improve learning efficiency. Such information has, for example, been exploited to guide preferential learning of individualized controller parameters~\cite{theiner2026prior_preferential}, while intermediate closed-loop observations can provide additional information beyond a single aggregate performance value~\cite{hirt2026time_series_closed_loop}. These results motivate the systematic exploitation of information that is already available from the control problem and its closed-loop execution.

A key observation underlying this work is that hierarchical decomposition of
the online control problem does not generally imply a separable
parameter-learning problem. Upper-level controllers generate plans or
references that become demands on lower controller levels, while their
realization feeds back through the physical system and affects subsequent
upper-level decisions. Closed-loop performance therefore depends jointly on
parameters distributed across the hierarchy. At the same time, the hierarchy
exposes informative quantities describing task execution, realization of
internally generated references, and control effort. Importantly, these cross-layer relations can themselves change with the
operating condition, since variations in the plant, environment, or task can
alter the closed-loop dynamics and thereby how demands and limitations
propagate through the controller hierarchy.
Hierarchical control thus provides three
closely related sources of structure for parameter learning: informative
performance components, cross-layer parameter coupling, and
context-dependent interactions. These motivate hierarchy-informed performance
modeling, cross-layer joint parameter learning, and context-dependent transfer
and adaptation, respectively.
Fig.~\ref{fig:method_overview} summarizes these three elements and their
interaction within the proposed framework.

Existing work addresses individual aspects of this problem. Joint
parameter tuning across interacting controller levels has been studied for
cascaded and hierarchical control architectures
~\cite{khosravi2022cascade_bo,rupenyan2021performance,
rupenyan2026joint_motion_tuning}, including robotic whole-body and locomotion
controllers~\cite{yuan2019whole_body_bo,li2023biped_tuning,
yeganegi2019robust_humanoid}. For example work on exoskeleton control similarly
showed the relevance of jointly learning parameters across controller
levels~\cite{hirt2025exoskeleton_joint}. Separately, composite and gray-box BO
exploit intermediate quantities or known problem structure instead of modeling
the complete objective as a single black box
~\cite{astudillo2019composite_bo,gonzalez2024bois}, while contextual BO has
been used to transfer controller-tuning information across changing operating
conditions~\cite{froehlich2022contextual_mpc,
widmer2023legged_safe_bo,xu2024violation_contextual,
le2025contextual_controller_adaptation}.
Related contextual policy-search approaches learn humanoid walking-controller
parameters that generalize across desired walking speeds
~\cite{abdolmaleki2016contextual_walking}.

We combine these ideas in a common contextual surrogate over the parameters
of all controller levels. Within the composite-BO formulation, the modeled
outputs are hierarchy-derived task, realization, and effort quantities with
known linear aggregation. Compared with scalar contextual BO, we retain these
component observations and their statistical dependence. For hierarchical
controller tuning, this complements joint parameter optimization with a shared
component-level model across operating conditions. This also distinguishes the
present approach from our previous hierarchical surrogate
~\cite{hirt2025hierarchical_surrogate}, which learns parameter-dependent
closed-loop dynamics. Here, we directly model informative episode-level
performance quantities, avoiding the need to identify and roll out a surrogate
of the complete closed-loop dynamics.

These three aspects arise from the same hierarchical closed-loop structure and
are therefore addressed within a single learning framework. We propose a
unified contextual Bayesian optimization framework for hierarchical controller
tuning that combines three complementary elements:
1) \emph{hierarchy-informed performance modeling}: we retain and jointly model
task performance, realization quality, and control effort instead of learning
only their scalar aggregate,
2) \emph{cross-layer joint parameter learning}: we optimize parameters across
interacting controller levels jointly, thereby retaining cross-layer
dependencies, and
3) \emph{context-dependent transfer and adaptation}: we condition the same
performance model on measurable operating conditions that affect the system
dynamics and, consequently, the interaction between controller levels. This
enables information sharing across related dynamical conditions and adaptation
to previously unseen operating conditions. Accordingly, the deployment goal is
to recommend parameters for the observed operating context, rather than one
controller that is fixed across all conditions.

We evaluate the framework for humanoid loco-manipulation using two-handed box pushing with a Unitree R1. A centroidal predictive controller generates motion and contact-wrench references that are realized by an inverse-dynamics whole-body controller, while box mass acts as the operating context.

In the following, Section~\ref{sec:fundamentals} introduces the control and learning
fundamentals. Section~\ref{sec:method} presents the proposed hierarchy-aware contextual parameter-learning framework. Section~\ref{sec:simulation} applies the method to hierarchical humanoid loco-manipulation control. We conclude in Section~\ref{sec:conclusion}.

\vspace{-1mm}
\section{Problem Formulation and Fundamentals}
\vspace{-1mm}
\label{sec:fundamentals}

We first formulate the closed-loop parameter-learning problem. Subsequently, we introduce parameterized model predictive control and whole-body control as two relevant optimization-based control frameworks, followed by Gaussian process regression and contextual Bayesian optimization.

\subsection{Problem Formulation}

We consider the nonlinear discrete-time dynamical system, with time index
$k\in\mathbb N_0$,
\begin{equation*}
    x_{k+1}=f(x_k,u_k;\chi),
\end{equation*}
where $f:\mathbb R^{n_{\mathrm{x}}}\times\mathbb R^{n_{\mathrm{u}}}
\times\mathcal X_{\upchi}\rightarrow\mathbb R^{n_{\mathrm{x}}}$,
$x_k\in\mathbb R^{n_{\mathrm{x}}}$ denotes the system state,
$u_k\in\mathbb R^{n_{\mathrm{u}}}$ denotes the control input, and
$\chi\in\mathcal X_{\upchi}\subset\mathbb R^{n_{\upchi}}$ is a measurable operating context that remains constant during one closed-loop experiment.

The system is controlled by a parameterized control architecture
\begin{equation*}
    u_k=\pi(x_k;\theta),
    \qquad
    \pi:\mathbb R^{n_{\mathrm{x}}}\times\Theta\rightarrow\mathbb R^{n_{\mathrm{u}}},
    \qquad
    \theta\in\Theta\subset\mathbb R^{n_{\uptheta}}.
\end{equation*}
For a fixed initial state $x_0$ and experiment horizon $K\in\mathbb N$,
the controller parameters $\theta$ and operating context $\chi$
determine the closed-loop state and input sequences
$\{x_k\}_{k=0}^{K}$ and $\{u_k\}_{k=0}^{K-1}$.
Their performance is evaluated by the scalar loss functional
$\mathcal L:(\mathbb R^{n_{\mathrm{x}}})^{K+1}\times
(\mathbb R^{n_{\mathrm{u}}})^K\rightarrow\mathbb R$ according to
\begin{equation}
    J(\theta,\chi)
    =
    \mathcal L
    \left(
        \{x_k\}_{k=0}^{K},
        \{u_k\}_{k=0}^{K-1}
    \right).
    \label{eq:closed_loop_cost}
\end{equation}
Our objective is to determine
\begin{equation*}
    \theta^\star(\chi)
    =
    \arg\min_{\theta\in\Theta}
    J(\theta,\chi).
\end{equation*}
Note that each evaluation of \eqref{eq:closed_loop_cost} requires a potentially
costly closed-loop experiment or simulation, while analytical gradients with
respect to $\theta$ are often  not available. This motivates the use of
Bayesian optimization as a sample-efficient derivative-free method for
closed-loop parameter learning.
In our application, model predictive control (MPC) and whole-body control (WBC)
form the upper and lower controllers in Fig.~\ref{fig:method_overview}, respectively.

\subsection{Parameterized Model Predictive Control}

MPC computes control inputs by repeated solution of a
finite-horizon optimal control problem~\cite{rawlings2017model}. For a prediction
horizon $N\in\mathbb N$, let $\hat x_{i|k}$ and $\hat u_{i|k}$ denote the state
and input predicted for time $k+i$ at time $k$, and let
$\mathbf{\hat u}_k=[\hat u_{0|k}^\top\ \cdots\ \hat u_{N-1|k}^\top]^\top$.
For a parameter vector $\theta_{\mathrm{MPC}}$, a general formulation is
\begin{equation}
\begin{aligned}
\min_{\mathbf{\hat u}_k}\quad
&
\sum_{i=0}^{N-1}
\ell_{\theta_{\mathrm{MPC}}}
\left(
    \hat x_{i|k},
    \hat u_{i|k}
\right)
+
V_{\theta_{\mathrm{MPC}}}
\left(
    \hat x_{N|k}
\right)
\\
\text{s.t.}\quad
&
\hat x_{0|k}=x_k,\\
&
\hat x_{i+1|k}
=
\hat f_{\theta_{\mathrm{MPC}}}
\left(
    \hat x_{i|k},
    \hat u_{i|k}
\right),
\qquad i=0,\ldots,N-1,\\
&
\hat x_{i|k}\in\mathcal X_{\theta_{\mathrm{MPC}}},
\qquad
\hat u_{i|k}\in\mathcal U,
\qquad i=0,\ldots,N-1,\\
&
 \hat x_{N|k}\in\mathcal X_{\theta_{\mathrm{MPC}}}.
\end{aligned}
\label{eq:mpc}
\end{equation}
Here, $\hat f_{\theta_{\mathrm{MPC}}}$ is the (parameterized) prediction model,
$\ell_{\theta_{\mathrm{MPC}}}$ and $V_{\theta_{\mathrm{MPC}}}$ are the (parameterized) stage and
terminal costs, and $\mathcal X_{\theta_{\mathrm{MPC}}}$ and $\mathcal U$ are the
admissible state and input sets. 
The applied control input is the first element of the optimized input sequence,
such that repeated solution of \eqref{eq:mpc} defines a parameterized feedback
policy $u_k=\pi_{\mathrm{MPC}}(x_k;\theta_{\mathrm{MPC}})$.
The upper controller in Fig.~\ref{fig:method_overview} uses a model predictive
path integral (MPPI) implementation of MPC and WBC realizes its plans.

\subsection{Whole-Body Control}

WBC, the lower controller in Fig.~\ref{fig:method_overview}, provides an optimization-based framework for
controlling floating-base articulated systems under contact and actuation
constraints. The rigid-body dynamics are given by
\begin{equation*}
M(q)\dot\nu+b(q,\nu)
=
S^\top\tau
+
J_{\mathrm{c}}(q)^\top w
+
Q_{\mathrm{e}},
\end{equation*}
where $q$ and $\nu$ denote generalized position and velocity, $M(q)$ is the
inertia matrix, $b(q,\nu)$ collects Coriolis and gravity terms, $S$ selects the
actuated joints, $\tau$ denotes joint torques, $J_{\mathrm{c}}(q)$ is the contact
Jacobian, $w$ denotes contact wrenches, and $Q_{\mathrm{e}}$ contains external
generalized forces.

For a parameter vector $\theta_{\mathrm{WBC}}$, a common inverse-dynamics WBC
formulation determines generalized accelerations and contact wrenches through
the weighted quadratic program
\begin{equation}
\begin{aligned}
\min_y\quad
&
\frac12 y^\top R_{\theta_{\mathrm{WBC}}} y
+
\frac12
\sum_{j\in\mathcal T}
\rho_{j,\theta_{\mathrm{WBC}}}
\left\|
A_j y-b_{j,\theta_{\mathrm{WBC}}}
\right\|_2^2
\\
\text{s.t.}\quad
&
A_{\mathrm{dyn}}y=b_{\mathrm{dyn}},\\
&
\underline d_{\theta_{\mathrm{WBC}}}
\le
C y
\le
\overline d_{\theta_{\mathrm{WBC}}}.
\end{aligned}
\label{eq:wbc_general}
\end{equation}
where $y=[\dot\nu^\top\ w^\top]^\top$ contains generalized accelerations and
contact wrenches, $R_{\theta_{\mathrm{WBC}}}\succeq0$ is a regularization matrix,
and $A_jy-b_{j,\theta_{\mathrm{WBC}}}$ is the residual of task $j\in\mathcal T$
with weight $\rho_{j,\theta_{\mathrm{WBC}}}\geq0$. The matrices and vectors
$A_{\mathrm{dyn}},b_{\mathrm{dyn}},C,\underline d_{\theta_{\mathrm{WBC}}}$, and
$\overline d_{\theta_{\mathrm{WBC}}}$ encode the rigid-body dynamics and the
contact, joint, and actuator limits.
The parameter vector $\theta_{\mathrm{WBC}}$ may therefore affect task weights,
tracking gains or references, regularization, and constraint bounds. Repeated
solution of \eqref{eq:wbc_general} defines a parameterized feedback law for the
joint torques.
Together, these optimization-based controller levels motivate a probabilistic
model for their joint closed-loop performance.

\subsection{Gaussian Process Regression}
Gaussian process (GP) regression provides a probabilistic model of an unknown
function $\varphi:\mathbb R^{n_{\mathrm{z}}}\rightarrow\mathbb R$. A GP
$
    g(z)\sim
    \mathcal{GP}
    \bigl(
        \mu(z),
        k(z,z')
    \bigr)
$
is specified by a mean function $\mu$ and covariance function $k$.

Given (noisy) observations
\begin{equation*}
    \mathcal D^{(n)}=\{(z^{(i)},y^{(i)})\}_{i=1}^{n},\
    y^{(i)}=\varphi(z^{(i)})+\varepsilon^{(i)},
\end{equation*}
with $\varepsilon^{(i)}\overset{\mathrm{iid}}{\sim}\mathcal N(0,\sigma_\varepsilon^2),$ GP regression yields the posterior
$
    g(z_\star)\mid\mathcal D^{(n)}
    \sim
    \mathcal N
    \bigl(
        \mu^{(n)}(z_\star),
        \bigl(\sigma^{(n)}(z_\star)\bigr)^2
    \bigr),
$
with posterior mean and posterior variance given by
\begin{align*}
    \mu^{(n)}(z_\star)
    &=
    \mu(z_\star)
    +
    k(z_\star,Z)
    K_{\mathrm{y}}^{-1}
    \left(
        y-\mu(Z)
    \right),\\
    \bigl(\sigma^{(n)}(z_\star)\bigr)^2
    &=
    k(z_\star,z_\star)
    -
    k(z_\star,Z)
    K_{\mathrm{y}}^{-1}
    k(Z,z_\star).
\end{align*}
Here, $Z=[z^{(1)}\ \cdots\ z^{(n)}]^\top\in\mathbb R^{n\times n_{\mathrm{z}}}$,
$y=[y^{(1)}\ \cdots\ y^{(n)}]^\top$, and
$\mu(Z)=[\mu(z^{(1)})\ \cdots\ \mu(z^{(n)})]^\top$. The matrix
$k(Z,Z)\in\mathbb R^{n\times n}$ has $(i,j)$ entry $k(z^{(i)},z^{(j)})$, while
$k(z_\star,Z)=[k(z_\star,z^{(1)})\ \cdots\ k(z_\star,z^{(n)})]$,
$k(Z,z_\star)=k(z_\star,Z)^\top$, and
$K_{\mathrm{y}}=k(Z,Z)+\sigma_\varepsilon^2I_n$, where $I_n$ is the $n\times n$ identity matrix.

The GP hyperparameters are commonly inferred through marginal-likelihood maximization~\cite{rasmussen2006gaussian}.
This posterior representation is the basis for the sequential decisions made by
Bayesian optimization.

\subsection{Contextual Bayesian Optimization}

Bayesian optimization is a sequential approach for optimizing expensive
black-box objectives such as \eqref{eq:closed_loop_cost}. We denote the input
to the surrogate model by $z$. In standard BO with fixed context $\chi$, the surrogate input is obtained
from the controller parameters through the map
$
    z=\zeta(\theta)=\theta.
$
After $n$ evaluations, the available data are
$
    \mathcal D^{(n)}
    =
    \left\{
        (z^{(i)},J^{(i)})
    \right\}_{i=1}^{n},
    J^{(i)}=J\!\left(\theta^{(i)},\chi\right).
$
A Gaussian process surrogate is conditioned on $\mathcal D^{(n)}$ and provides a
posterior mean $\mu_J^{(n)}(z)$ and uncertainty $\sigma_J^{(n)}(z)$ over the unknown
objective.

An acquisition function $\alpha^{(n)}(z)$ uses this posterior to determine where the
objective should be evaluated next, balancing promising predicted performance
with uncertainty in unexplored regions. In standard BO, the next parameter
vector is selected according to
\begin{equation*}
    \theta^{(n+1)}
    =
    \arg\max_{\theta\in\Theta}
    \alpha^{(n)}(\theta).
\end{equation*}
The selected parameters are then applied to the closed-loop system, yielding
the observation $J^{(n+1)}$. This observation is appended to the
data set, after which the surrogate is updated and the procedure is repeated.

In contextual BO, the operating condition is observed before selecting the
controller parameters and is included in the surrogate input through
$
    z
    =
    \zeta(\theta,\chi)
    =
    \begin{bmatrix}
        \theta^\top &
        \chi^\top
    \end{bmatrix}^\top.
$
The next parameter vector is selected by
\begin{equation}
    \theta^{(n+1)}
    =
    \arg\max_{\theta\in\Theta}
    \alpha^{(n)}\!\left(\zeta(\theta,\chi^{(n+1)})\right).
    \label{eq:contextual_selection}
\end{equation}
After evaluation, the corresponding input
$z^{(n+1)}=\zeta(\theta^{(n+1)},\chi^{(n+1)})$ and observed performance
$J^{(n+1)}=J(\theta^{(n+1)},\chi^{(n+1)})$ are added to the data set. Consequently,
observations collected under different operating conditions update one shared
performance model, while only the controller parameters are optimized and the
context remains fixed to the observed operating condition.
These fundamentals provide the notation needed to formulate the proposed
hierarchy-aware contextual learning method.

\vspace{-1mm}
\section{Proposed Method}
\vspace{-1mm}
\label{sec:method}

The following subsections introduce joint cross-layer parameter learning,
hierarchy-informed performance modeling, and contextual transfer and adaptation.

\subsection{Cross-Layer Joint Parameter Learning}

The complete parameter vector is written as
\begin{equation*}
    \theta
    =
    \begin{bmatrix}
        \theta_1^\top&
        \cdots&
        \theta_L^\top
    \end{bmatrix}^\top,
\end{equation*}
where the blocks correspond to parameters associated with different controller levels or internal reference generators.

For a prescribed context $\chi^{(n+1)}$, the next controller parameters are
selected jointly according to the contextual acquisition rule
\eqref{eq:contextual_selection}.
In contrast to sequential tuning, no controller block is fixed while another
is optimized. Favorable and unfavorable cross-layer parameter combinations
therefore directly update the same performance model. We next describe the
hierarchy-informed surrogate that supplies this joint acquisition.

\subsection{Hierarchy-Informed Performance Modeling}

We assume that the closed-loop objective can be evaluated from $n_{\mathrm{J}}$
observable performance components,
\begin{equation*}
    \mathbf J(\theta,\chi)
    =
    \begin{bmatrix}
        J_1(\theta,\chi)&
        \cdots&
        J_{n_{\mathrm{J}}}(\theta,\chi)
    \end{bmatrix}^\top,
\end{equation*}
with known scalar aggregation
\begin{equation}
    J(\theta,\chi)
    =
    a^\top
    \mathbf J(\theta,\chi),
    \qquad
    a\in\mathbb R_+^{n_{\mathrm{J}}}.
    \label{eq:component_aggregation}
\end{equation}

Performance components are constructed by following the information flow
through the controller hierarchy: external task achievement, realization of
references passed between controller levels, and the actuation effort required
to realize them. The hierarchy thus informs the observed performance quantities
and parameter grouping, while the surrogate learns their statistical
relationships. Importantly, this decomposition concerns the performance
measure and does not imply separable parameter dependence.

We jointly model the performance components with the input
$z=[\theta^\top\ \chi^\top]^\top$ and a multi-output GP covariance~\cite{bonilla2007multitask}
\begin{equation*}
    \operatorname{Cov}
    \left[
        g_j(z),
        g_{j'}(z')
    \right]
    =
    B_{jj'}
    k(z,z'),
\end{equation*}
where $g_j$ models component $J_j$ and $B\succeq0$ is the
$n_{\mathrm{J}}\times n_{\mathrm{J}}$ output covariance matrix. Kernel
hyperparameters, the output covariance matrix $B$, and component-specific
observation-noise variances are estimated jointly by maximizing the log marginal
likelihood of the component observations with hyperparameter priors.

Let $\boldsymbol\mu^{(n)}(z)\in\mathbb R^{n_{\mathrm{J}}}$ and
$\Sigma^{(n)}(z)\in\mathbb R^{n_{\mathrm{J}}\times n_{\mathrm{J}}}$ denote the
posterior mean and covariance of the component vector. Since
\eqref{eq:component_aggregation} is known and linear, the posterior of the
scalar objective follows directly as
\begin{align*}
    \mu_J^{(n)}(z)
    &=
    a^\top
    \boldsymbol\mu^{(n)}(z),\\
    \bigl(\sigma_J^{(n)}(z)\bigr)^2
    &=
    a^\top
    \Sigma^{(n)}(z)
    a.
\end{align*}
This scalar posterior supplies the acquisition used for joint parameter
selection and retains the component-level information. We next use the context
input to transfer these learned relations across operating conditions.

\subsection{Context-Dependent Transfer and Adaptation}

Contextual modeling provides a shared representation of hierarchical controller
performance across changing operating conditions. Its purpose is not only to
pool data from related experiments, but to support context-dependent deployment:
after observing the current context, the controller parameters can be tailored
to that condition. Let $\mathcal X_{\mathrm{tr}}$ denote a finite set of training
contexts and let $\chi_{\mathrm{new}}\in\mathcal X_{\upchi}\setminus
\mathcal X_{\mathrm{tr}}$ denote a previously unseen deployment context. Once
the surrogate has been trained, the recommendation for any observed context is
given by
\begin{equation}
    \widehat\theta^{(n)}(\chi)
    =
    \arg\min_{\theta\in\Theta}
    \mu_J^{(n)}\!\left(\zeta(\theta,\chi)\right).
    \label{eq:contextual_rec}
\end{equation}

Equation~\eqref{eq:contextual_rec} selects the parameter vector with the
smallest predicted aggregate cost at the observed context. Because the same
contextual surrogate is fitted to all observations, it transfers
information across operating conditions while allowing the recommendation to
vary with $\chi$.

At the unseen deployment context $\chi_{\mathrm{new}}$, the zero-shot
recommendation $\widehat\theta^{(n)}(\chi_{\mathrm{new}})$ is obtained from the
context-labeled training observations according to \eqref{eq:contextual_rec}.
Additional local experiments are then appended to the data set and update the
same surrogate, yielding few-shot adaptation of the recommendation at
$\chi_{\mathrm{new}}$. Thus, the hierarchy-informed performance model supplies
a prior for the new operating condition before any local measurements are
collected.

We now instantiate this framework for the hierarchical humanoid
loco-manipulation system.

\vspace{-1mm}
\section{Simulation Results}
\vspace{-1mm}
\label{sec:simulation}

We evaluate the proposed framework on a hierarchical humanoid
loco-manipulation problem. A Unitree R1 humanoid performs two-handed
physical box pushing while walking. A centroidal predictive controller
and an inverse-dynamics whole-body controller are combined into one
hierarchical control architecture, providing a representative setting in
which parameters distributed across multiple controller levels jointly
determine the closed-loop behavior.

\vspace{-1mm}
\subsection{Humanoid Loco-Manipulation Setup}
\vspace{-1mm}

\begin{figure}[t]
    \centering
    \includegraphics[width=.8\columnwidth]{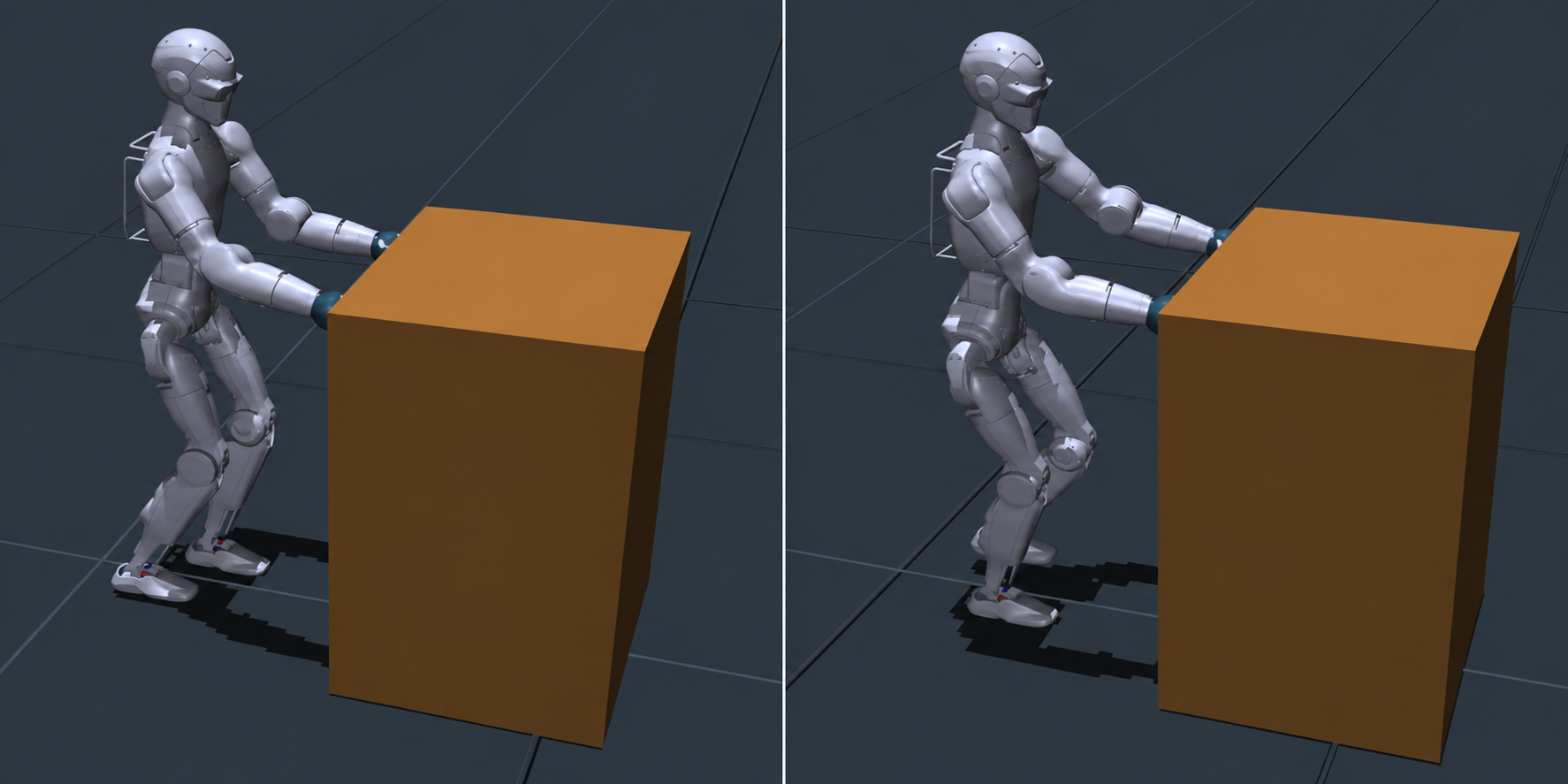}
    \par\vspace{-2mm}
    \caption{The simulated humanoid box-pushing task.}
    \vspace{-2mm}
    \label{fig:box_pushing_sequence}
\end{figure}

In MuJoCo, a Unitree R1 humanoid walks while pushing a free rigid box through
two-handed contact. Every method is evaluated on the same $20\,\mathrm s$
external task, specified by a smooth desired longitudinal box-velocity profile.
The first $2\,\mathrm{s}$ are reserved for contact preparation and performance is
scored over the remaining $18\,\mathrm{s}$.
The box is not included in the predictive model; instead, measured hand
reactions communicate its influence to both controller levels. Two stages of a
representative rollout are shown in Fig.~\ref{fig:box_pushing_sequence}.

The external task remains fixed, while tunable gains and temporal shifts allow
the internal gait and hand-motion references to respond differently to it.
Consequently, BO-based parameter learning changes how locomotion and manipulation are coordinated
through the hierarchy without changing the task used for evaluation. We next
describe the two controller levels and the parameterization.

\vspace{-1mm}
\subsection{Hierarchical Robot Controller}
\vspace{-1mm}

The upper level is a centroidal Model Predictive Path Integral (MPPI)
controller~\cite{williams2017mppi}. It plans center-of-mass (CoM) and momentum motion together with supporting foot
wrenches and incorporates the measured box-interaction wrench. The lower level
is the inverse-dynamics WBC introduced in \eqref{eq:wbc_general}, which realizes
the centroidal state and wrench plan on the full articulated robot while
enforcing dynamics, contact, joint, and actuator constraints.

Simulation and WBC updates run at $500\,\mathrm{Hz}$, with a
$2\,\mathrm{ms}$ simulation timestep. The centroidal MPPI controller updates
at $50\,\mathrm{Hz}$, using 512 sampled trajectories with exponential cost
weighting over a $1\,\mathrm{s}$ prediction horizon with $20\,\mathrm{ms}$
discretization.

We optimize the combined parameter vector
$\theta=[\theta_{\mathrm{MPPI}}^\top\;
\theta_{\mathrm{WBC}}^\top\;
\theta_{\mathrm{ref}}^\top\;
\theta_{\mathrm{hand}}^\top\;
\theta_{\mathrm{cpl}}^\top]^\top\in\mathbb R^{28}$.
The five parameter groups are summarized in
Table~\ref{tab:parameter_groups}. Here, $\theta_{\mathrm{MPPI}}$ contains
parameters of the predictive controller, $\theta_{\mathrm{WBC}}$ parameters
of the whole-body controller, $\theta_{\mathrm{ref}}$ internal motion
references, $\theta_{\mathrm{hand}}$ parameters of the hand-tracking task, and
$\theta_{\mathrm{cpl}}$ parameters governing the response between the external
task and the internal references.

\begin{table}[t]
\centering
\caption{Optimized parameter groups.}
\label{tab:parameter_groups}
\scriptsize
\begin{tabular}{lll}
\toprule
Group & Optimized quantities & Dim.\\
\midrule
$\theta_{\mathrm{MPPI}}$
& MPPI temperature; motion, momentum, effort, & 6\\[-1mm]
& and wrench-variation weights & \\

$\theta_{\mathrm{WBC}}$
& CoM, torso, foot, posture, momentum, and & 11\\[-1mm]
& wrench-tracking gains and task weights & \\

$\theta_{\mathrm{ref}}$
& Torso pitch, CoM offsets, step duration, & 5\\[-1mm]
& and swing clearance & \\

$\theta_{\mathrm{hand}}$
& Hand-tracking gain and task weight & 2\\

$\theta_{\mathrm{cpl}}$
& Hand gain $\kappa_{\mathrm h}$, hand delay $\tau_{\mathrm h}$, & 4\\[-1mm]
& gait gain $\beta$, gait delay $\tau_{\mathrm g}$ & \\
\bottomrule
\end{tabular}
\vspace{-6mm}
\end{table}

The response parameters $\beta$ and $\tau_{\mathrm g}$ scale and temporally
shift how the internal gait reference reacts to the external pushing task,
whereas $\kappa_{\mathrm h}$ and $\tau_{\mathrm h}$ scale and shift the coordination of the
hand-motion reference with the gait. The remaining internal references specify
the desired posture and stepping behavior.

The two controller levels are coupled through the planned centroidal motion and
contact wrenches passed from the predictive controller to the WBC, and through
the physical closed loop: upper-level choices change the demands placed on the
WBC, while their realization affects subsequent planning. For the sequential
baselines only, the parameter vector is divided into a
ten-dimensional upper-level block $\theta_1$, containing the six predictive-controller
parameters, two CoM references, and the two gait-response parameters, and an
18-dimensional lower-level block $\theta_2$ containing the remaining parameters.
The sequential orders are denoted by $1\!\rightarrow\!2$ and
$2\!\rightarrow\!1$.
The proposed method instead optimizes the complete parameter vector jointly.

\vspace{-1mm}
\subsection{Hierarchy-Informed Performance Components}
\vspace{-1mm}

Every closed-loop experiment provides the three performance components
\begin{equation*}
    J = J_{\mathrm T}+J_{\mathrm R}+J_{\mathrm E},
\end{equation*}
with equal aggregation weights applied to the normalized components. The components quantify external task
execution ($J_{\mathrm T}$), realization of the internally generated motion ($J_{\mathrm R}$), and actuation
effort ($J_{\mathrm E}$), respectively.

For the task component, the scored part of the episode is divided into
$N_{\mathrm w}$ fixed time windows with boundaries $t_j$. Let
$\Delta p_j=x_{\mathrm{box}}(t_{j+1})-x_{\mathrm{box}}(t_j)$ denote the
actual forward box displacement in window $j$, and let
$\Delta p_j^{\mathrm d}$ denote the corresponding displacement prescribed by
the fixed external reference. Furthermore, $y_{\mathrm{box}}$ denotes the
lateral box position and $y_{\mathrm{box},0}$ its initial value. Using
$s_p=s_y=0.05\,\mathrm m$, the three components are compactly written as
\begin{align*}
J_{\mathrm T}
&=
\frac{1}{N_{\mathrm w}}
\sum_{j=1}^{N_{\mathrm w}}
\left(
    \frac{\Delta p_j-\Delta p_j^{\mathrm d}}{s_p}
\right)^2
+
\left\langle
\left(
    \frac{y_{\mathrm{box}}-y_{\mathrm{box},0}}{s_y}
\right)^2
\right\rangle ,
\\
J_{\mathrm R}
&=
\frac{1}{2}
\left(
    \frac{\langle r_{\mathrm R}\rangle}{s_{\mathrm R}}
    +
    \langle e_{\mathrm h}\rangle
\right),
\\
J_{\mathrm E}
&=
\frac{1}{s_{\mathrm E}}
\left\langle
    \|\widetilde{\tau}\|_{\mathrm{av}}^2
    +
    0.1\,
    \|0.02\,\dot{\widetilde{\tau}}\|_{\mathrm{av}}^2
\right\rangle .
\end{align*}
Here, $\langle\cdot\rangle$ denotes the temporal average over the scored
portion of the experiment. The quantity $r_{\mathrm R}$ is the mean normalized
squared tracking error of CoM position and velocity, angular momentum, and foot
position and velocity relative to the centroidal plan. The quantity
$e_{\mathrm h}$ is the mean squared hand-position tracking error. $s_{\mathrm R}$ and $s_{\mathrm E}$ are fixed normalization constants.

For the effort component, $\tau$ denotes the vector of joint torques,
$\widetilde{\tau}_i=\tau_i/\tau_i^{\max}$ is the torque of joint $i$
normalized by its torque limit, and $\dot{\widetilde{\tau}}$ denotes its
time derivative. We use
$\|v\|_{\mathrm{av}}^2=\|v\|_2^2/\dim(v)$ to denote the mean squared value
across the elements of a vector $v$.

The progress-based task component $J_{\mathrm T}$ evaluates whether the box
achieves the required forward progress within each time interval while
remaining on the desired straight path. In contrast to requiring the box
velocity to track a continuously prescribed reference at every instant, this
provides temporal freedom in how the controller realizes the required progress.
Importantly, the external box reference is fixed independently of the internal
gait and hand references, such that BO cannot improve the task score by
changing the task itself.

All three components are external evaluation quantities and are distinct from
the objectives optimized internally by MPPI and WBC. Together, they distinguish
task achievement, realization of the generated motion, and the actuation
required to produce it.

\vspace{-1mm}
\subsection{Bayesian Optimization Setup}
\vspace{-1mm}
\label{sec:bo_setup}

Box mass is the scalar context, $\chi=m_{\mathrm{box}}$, with training set
$\mathcal X_{\mathrm{tr}}=\{2,3,5,6\}\,\mathrm{kg}$. Together with the 28
controller parameters, the contextual surrogate has 29 inputs. Each
closed-loop evaluation lasts $20\,\mathrm s$.

We compare the proposed correlated component model with independent component
GPs, contextual and context-free scalar BO, both orders of sequential
upper-/lower-level tuning, and random search. Every method receives 48
closed-loop evaluations and is repeated for ten independent seeds.
All methods use identical parameter bounds. The sequential
baselines freeze the first optimized parameter block before tuning the second.
All methods share the same initial parameter vector, evaluated once at each
training mass; these four initial rollouts count toward the 48-evaluation budget.
Sequential methods switch parameter blocks after 32 evaluations, including
initialization, and use the remaining 16 evaluations to tune the second block.
After initialization, contextual methods cycle through the training masses in
the fixed order $2,3,5,6\,\mathrm{kg}$.

The context-free universal baseline seeks one parameter vector for all training
contexts. Each candidate is evaluated at every training mass, and a scalar GP
without a context input models the mean observed cost across these masses. Each
rollout counts separately toward the common evaluation budget. This baseline
represents universal controller tuning, whereas the contextual methods recommend
parameters conditioned on the observed mass.

Failed trials, including premature termination, insufficient task progress,
loss of sustained contact, or excessive path/tilt deviation, count toward the
evaluation budget but are excluded from the performance surrogate. All BO
variants use the same feasibility treatment. An auxiliary GP regresses binary
eligibility labels, with posterior mean $\mu_{\mathrm f}^{(n)}(z)$ and variance
$\bigl(\sigma_{\mathrm f}^{(n)}(z)\bigr)^2$ on the label scale. We approximate the probability
of an eligible rollout by the Gaussian threshold probability
\begin{equation*}
    p_{\mathrm f}^{(n)}(z)
    =\Phi\!\left(
        \frac{\mu_{\mathrm f}^{(n)}(z)-0.5}
        {\sqrt{\bigl(\sigma_{\mathrm f}^{(n)}(z)\bigr)^2+0.01}}
    \right),
\end{equation*}
where $\Phi$ is the standard normal cumulative distribution function and $z$
denotes the input of the corresponding surrogate. When all observations are
eligible, $p_{\mathrm f}^{(n)}(z)=1$.

We use analytic expected improvement (EI) of the scalar objective posterior,
weighted by estimated feasibility, as $\widetilde\alpha^{(n)}$ in place of
$\alpha^{(n)}$ in \eqref{eq:contextual_selection},
\begin{equation*}
    \widetilde\alpha^{(n)}(z)
    =p_{\mathrm f}^{(n)}(z)\,\operatorname{EI}^{(n)}(z).
\end{equation*}
EI is computed relative to the best eligible observed cost at the queried
context; the universal baseline instead uses the best eligible average cost
across training contexts.
Final recommendations minimize predicted cost among candidates satisfying
$p_{\mathrm f}^{(n)}(z)\geq0.8$. If no candidate satisfies this threshold, the
candidate with the highest estimated feasibility is selected.

\vspace{-1mm}
\subsection{Learning and Adaptation Results}
\vspace{-1mm}

Since the globally optimal closed-loop cost is unknown, optimization
performance is evaluated using empirical simple regret. For each context
$\chi$, we define $J_{\mathrm{ref}}(\chi)$ retrospectively as the smallest
eligible cost observed across all methods and BO seeds over the complete
experiment. In contrast, $J_{\mathrm{best}}^{(n)}(\chi)$ denotes the smallest
eligible cost found within the respective optimization run up to evaluation
$n$. The resulting training regret is
\begin{equation}
    R_{\mathrm{emp}}^{(n)}
    =
    \frac{1}{|\mathcal X_{\mathrm{tr}}|}
    \sum_{\chi\in\mathcal X_{\mathrm{tr}}}
    \left(
        J_{\mathrm{best}}^{(n)}(\chi)
        -
        J_{\mathrm{ref}}(\chi)
    \right).
    \label{eq:empirical_regret}
\end{equation}
Failed evaluations count toward the evaluation budget but do not replace the
best eligible controller found so far. Hence,
\eqref{eq:empirical_regret} measures the gap between the best controller found
so far in a given optimization run and the best eligible controller observed
overall in retrospect, averaged over the training contexts.

\begin{figure*}[t]
    \centering
    \definecolor{boCorrelated}{HTML}{2B6F9D}
    \definecolor{boIndependent}{HTML}{7AAECF}
    \definecolor{boScalar}{HTML}{555555}
    \definecolor{boUniversal}{HTML}{999999}
    \definecolor{boSequentialHm}{HTML}{C09224}
    \definecolor{boSequentialMh}{HTML}{D9B85E}
    \definecolor{boRandom}{HTML}{8E44AD}
    \begin{minipage}[t]{0.49\textwidth}
        \centering
        \input{figures/training_regret.pgf}
    \end{minipage}
    \hfill
    \begin{minipage}[t]{0.49\textwidth}
        \centering
\begingroup
\providecommand{\plotfont}{\footnotesize}
\providecommand{\plotlegendfont}{\scriptsize}
\pgfplotsset{compat=1.18}
\definecolor{boCorrelated}{HTML}{2B6F9D}
\definecolor{boIndependent}{HTML}{7AAECF}
\definecolor{boScalar}{HTML}{555555}
\definecolor{boUniversal}{HTML}{999999}
\definecolor{boSequentialHm}{HTML}{C09224}
\definecolor{boSequentialMh}{HTML}{D9B85E}
\definecolor{boRandom}{HTML}{8E44AD}
\begin{tikzpicture}
\begin{axis}[
  width=\linewidth, height=0.57\linewidth,
  font=\plotfont, label style={font=\plotfont}, tick label style={font=\plotfont},
  axis line style={draw=black!60, line width=0.35pt},
  tick style={draw=black!60, line width=0.35pt},
  axis x line*=bottom, axis y line*=left,
  ymajorgrids, grid style={draw=black!10, line width=0.3pt},
  xlabel={Additional closed-loop evaluations},
  ylabel={Empirical regret},
  xmin=0.9, xmax=4.15, xtick={1,2,3,4},
  ymin=-0.0762635754296, ymax=1.27430101428,
  clip mode=individual,
]
\addplot[draw=none, const plot, name path=correlatedUpper] coordinates {(1,0.518524880045) (2,0.277868231082) (3,0.267532019307) (4,0.262126800151)};
\addplot[draw=none, const plot, name path=correlatedLower] coordinates {(1,0.0069989401628) (2,0.096464853725) (3,0.0903763101823) (4,0.0865870018517)};
\addplot[draw=none, fill=boCorrelated, fill opacity=0.10] fill between[of=correlatedUpper and correlatedLower];
\addplot[draw=none, const plot, name path=independentUpper] coordinates {(1,0.41161976531) (2,0.381051770937) (3,0.375878151849) (4,0.352369620469)};
\addplot[draw=none, const plot, name path=independentLower] coordinates {(1,0.164266604951) (2,0.133090655997) (3,0.117607699827) (4,0.0982210821845)};
\addplot[draw=none, fill=boIndependent, fill opacity=0.10] fill between[of=independentUpper and independentLower];
\addplot[draw=none, const plot, name path=scalarUpper] coordinates {(1,0.986159056385) (2,0.500654672858) (3,0.510287400698) (4,0.444207968811)};
\addplot[draw=none, const plot, name path=scalarLower] coordinates {(1,-0.0148742758972) (2,0.186469962551) (3,0.188504232766) (4,0.177137475491)};
\addplot[draw=none, fill=boScalar, fill opacity=0.10] fill between[of=scalarUpper and scalarLower];
\addplot[draw=none, const plot, name path=universalUpper] coordinates {(1,1.04045667312) (2,1.03917406016) (3,1.00599357327) (4,0.987822384248)};
\addplot[draw=none, const plot, name path=universalLower] coordinates {(1,0.777324991005) (2,0.753515736377) (3,0.74706450567) (4,0.744316115204)};
\addplot[draw=none, fill=boUniversal, fill opacity=0.10] fill between[of=universalUpper and universalLower];
\addplot[draw=none, const plot, name path=sequential_hmUpper] coordinates {(1,0.618628890007) (2,0.549366876554) (3,0.38490051373) (4,0.379412394317)};
\addplot[draw=none, const plot, name path=sequential_hmLower] coordinates {(1,0.137886892185) (2,0.136183796026) (3,0.128571656755) (4,0.117879934012)};
\addplot[draw=none, fill=boSequentialHm, fill opacity=0.10] fill between[of=sequential_hmUpper and sequential_hmLower];
\addplot[draw=none, const plot, name path=sequential_mhUpper] coordinates {(1,0.519507411464) (2,0.487009675526) (3,0.460563488202) (4,0.402742889569)};
\addplot[draw=none, const plot, name path=sequential_mhLower] coordinates {(1,0.18337286782) (2,0.163737888984) (3,0.0899491651248) (4,0.08130533141)};
\addplot[draw=none, fill=boSequentialMh, fill opacity=0.10] fill between[of=sequential_mhUpper and sequential_mhLower];
\addplot[draw=none, const plot, name path=randomUpper] coordinates {(3,1.21291171475) (4,1.19533732176)};
\addplot[draw=none, const plot, name path=randomLower] coordinates {(3,0.946265959927) (4,0.892855087354)};
\addplot[draw=none, fill=boRandom, fill opacity=0.10] fill between[of=randomUpper and randomLower];
\addplot[color=boCorrelated, solid, line width=0.8pt, const plot, mark=*, mark size=1.5pt, mark repeat=1] coordinates {(1,0.262761910104) (2,0.187166542404) (3,0.178954164745) (4,0.174356901001)};
\addplot[color=boIndependent, dashed, line width=0.8pt, const plot, mark=triangle*, mark size=1.5pt, mark repeat=1] coordinates {(1,0.287943185131) (2,0.257071213467) (3,0.246742925838) (4,0.225295351327)};
\addplot[color=boScalar, solid, line width=0.8pt, const plot, mark=square*, mark size=1.5pt, mark repeat=1] coordinates {(1,0.485642390244) (2,0.343562317705) (3,0.349395816732) (4,0.310672722151)};
\addplot[color=boUniversal, dotted, line width=0.8pt, const plot, mark=diamond*, mark size=1.5pt, mark repeat=1] coordinates {(1,0.908890832061) (2,0.896344898268) (3,0.876529039472) (4,0.866069249726)};
\addplot[color=boSequentialHm, dashdotted, line width=0.8pt, const plot, mark=triangle*, mark size=1.5pt, mark repeat=1] coordinates {(1,0.378257891096) (2,0.34277533629) (3,0.256736085243) (4,0.248646164164)};
\addplot[color=boSequentialMh, dashed, line width=0.8pt, const plot, mark=x, mark size=1.5pt, mark repeat=1] coordinates {(1,0.351440139642) (2,0.325373782255) (3,0.275256326663) (4,0.242024110489)};
\addplot[color=boRandom, solid, line width=0.8pt, const plot, mark=pentagon*, mark size=1.5pt, mark repeat=1] coordinates {(3,1.07958883734) (4,1.04409620456)};
\addplot[black!45, line width=0.4pt] coordinates {(0,0) (48,0)};
\end{axis}
\end{tikzpicture}
\endgroup
    \end{minipage}
    \vspace{-1mm}
    \par\ref{boMethodLegend}
    \vspace{-5mm}
    \caption{Empirical regret during training (left)
    and adaptation at the unseen $4\,\mathrm{kg}$ mass (right). Legend
    labels indicate, in order, context use ($c$), joint cross-layer tuning
    ($j$), and component-wise performance observations ($p$). $\mathrm{x}$ denotes an
    absent feature. Curves show the mean, and shaded regions indicate one standard deviation across ten independent runs.}
    \label{fig:optimization_results}
    \vspace{-7mm}
\end{figure*}

The left panel of Fig.~\ref{fig:optimization_results} shows the optimization
performance over the 48 training evaluations. The proposed
correlated component model exhibits the fastest decrease in mean regret
over most of the experiment and obtains the smallest final value,
$0.196\pm0.107$. The independent-component model reaches
$0.270\pm0.135$, whereas contextual scalar BO reaches
$0.342\pm0.133$. Thus, retaining the component-wise observations yielded
lower mean regret compared with immediately reducing each experiment to its
scalar sum, while additionally retaining their statistical dependence
results in the lowest final mean regret. We therefore interpret this
comparison as evidence for the complete hierarchy-informed surrogate rather
than attributing the difference solely to output correlation.

The comparison with the two sequential methods provides a more nuanced
picture of cross-layer joint parameter learning. Sequential
$2\!\rightarrow\!1$ and $1\!\rightarrow\!2$ reach final regrets of
$0.270\pm0.076$ and $0.292\pm0.109$, respectively, compared with
$0.342\pm0.133$ for joint contextual scalar BO. Consequently, the
results do not indicate that searching the complete parameter space
jointly is by itself more sample efficient than exploiting the
hierarchical partition sequentially. The sequential structure can serve
as an effective dimensionality reduction under a limited experiment
budget. At the same time, the best overall result is obtained by the
proposed method, which retains cross-layer parameter combinations while
using the additional information contained in the component-wise
performance observations. This supports treating joint parameter
learning and hierarchy-informed performance modeling as complementary
parts of the proposed framework rather than independent improvements.

The effect of contextual learning is particularly clear. Contextual
scalar BO reaches $0.342\pm0.133$, whereas the otherwise scalar,
context-free universal model reaches only $0.805\pm0.104$. Random
search ends at $0.662\pm0.103$. The universal model optimizes a single
parameter vector based on its average performance across all training
masses and therefore favors compromise settings, whereas the empirical
regret is evaluated from the best controller found at each context.
This helps explain why even random search can outperform the universal
baseline when favorable parameters differ across operating conditions.
The large difference between contextual and universal scalar BO indicates
that averaging observations across box masses obscures systematic changes
of the closed-loop performance landscape.

To evaluate transfer to a new operating condition, all methods are
subsequently applied at the previously unseen mass
$
    \chi_{\mathrm{new}}
    =
    4\,\mathrm{kg}.
$
Each method is allowed four additional closed-loop experiments. The first
adaptation experiment evaluates the controller recommended by the
pretrained model without any local data and therefore constitutes the
measured zero-shot performance. Subsequent experiments update the model
using observations at $4\,\mathrm{kg}$.

The right panel of Fig.~\ref{fig:optimization_results} shows the resulting few-shot
behavior. The correlated model already obtains a mean empirical regret of
$0.263\pm0.256$ in the first evaluation and improves to
$0.174\pm0.088$ after four local experiments. The independent-component
model improves from $0.288\pm0.124$ to $0.225\pm0.127$, while
contextual scalar BO reaches $0.311\pm0.134$ after four evaluations.
The sequential methods also transfer effectively, reaching
$0.249\pm0.131$ for $1\!\rightarrow\!2$ and
$0.242\pm0.161$ for $2\!\rightarrow\!1$.

In contrast, the context-free universal model transfers poorly, with its
empirical regret decreasing only from $0.909\pm0.132$ to
$0.866\pm0.122$. Random search rarely identifies an eligible controller
within the four available trials. This is also reflected in the 40 few-shot experiments per method. Specifically, the number of failures is 3 for the proposed method, 2 for
the independent-component model, 3 for contextual scalar BO, and 4 for
sequential $2\!\rightarrow\!1$, versus 11 for sequential $1\!\rightarrow\!2$,
18 for the universal model, and 35 for random search. The contextual models
therefore provide a substantially more informative starting point at the
unseen operating condition. Among them, the proposed correlated component
model yields both the smallest measured zero-shot regret and the smallest
regret after four adaptation experiments.

Overall, the ablations indicate that the three sources of structure
considered in this work are complementary. Explicit context provides a
clear benefit over pooling different dynamical conditions into a single
performance function. The sequential baselines show that the known
controller hierarchy itself can be exploited to reduce the effective
search dimension and that joint optimization alone does not guarantee
faster learning. The lowest training and adaptation regret is obtained
when the complete parameterization is retained while the BO surrogate
additionally exploits the component-wise observations and the operating
context.

\begin{figure}[t]
    \centering
    \input{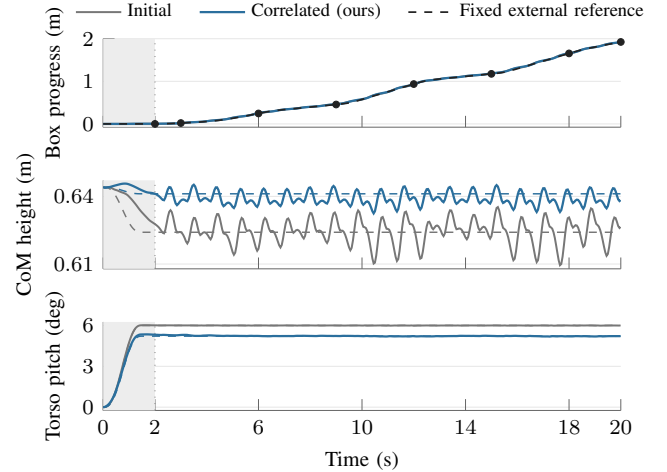}
    \vspace{-9mm}
    \caption{Representative closed-loop trajectories at
    $2\,\mathrm{kg}$. The top panel shows the identical box-progress task for
    both controllers under a shared external reference (dots). The middle and bottom
    panels show center-of-mass height and torso pitch, respectively, together with each
    controller's tunable reference. Solid lines denote realized trajectories,
    dashed lines denote references.}
    \label{fig:closed_loop_trajectories}
    \vspace{-7mm}
\end{figure}

For qualitative insight into the resulting controller behavior,
Fig.~\ref{fig:closed_loop_trajectories} compares the predefined initial
controller with a controller obtained using the proposed method under the
same $2\,\mathrm{kg}$ conditions. Both controllers are evaluated against
the same external box trajectory.
As shown in Fig.~\ref{fig:closed_loop_trajectories}, both controllers
achieve essentially the same box progress, while the optimized
controller adopts a higher CoM trajectory with more pronounced oscillations and a slightly reduced torso pitch. These
differences reflect a substantially different internal realization of the
same external pushing task. The figure therefore illustrates how parameters
distributed across the hierarchy modify the internal closed-loop behavior
while the external task remains fixed.

\section{Conclusions}
\vspace{-1mm}
\label{sec:conclusion}

We proposed a hierarchy-aware contextual Bayesian optimization framework for parameter learning in hierarchical control systems. The three elements of the method arise from the same closed-loop hierarchical structure: hierarchy-informed performance modeling exploits informative quantities exposed by the controller hierarchy, cross-layer joint parameter learning retains interactions between parameters of different controller levels, and context-dependent transfer captures how these coupled performance relations change with operating conditions. The multi-output surrogate preserves the covariance between informative performance components while their known aggregation defines the overall objective.

The framework was instantiated for hierarchical humanoid loco-manipulation control, jointly tuning centroidal predictive control, whole-body control, and internal motion references during physical box pushing. The proposed approach achieved the best overall performance during both training and few-shot adaptation.
Future work will investigate additional operating conditions, more complex loco-manipulation tasks, and transfer to robotic hardware.

\vspace{-2mm}
\bibliographystyle{ieeetr}
\AddToHook{cmd/thebibliography/after}{\vspace{-1mm}}
\bibliography{bibliography}

\end{document}